\documentclass[conference]{IEEEtran}
\IEEEoverridecommandlockouts
\usepackage{cite}
\usepackage{amsmath,amssymb,amsfonts}
\usepackage{algorithmic}
\usepackage{graphicx}
\usepackage{textcomp}
\usepackage{xcolor}
\usepackage{url}
\usepackage{cite}
\usepackage[hidelinks]{hyperref}
\usepackage{multirow}
\usepackage{acronym}
\usepackage{float}

\def\BibTeX{{\rm B\kern-.05em{\sc i\kern-.025em b}\kern-.08em
    T\kern-.1667em\lower.7ex\hbox{E}\kern-.125emX}}
\begin{document}

% -----------------------------------------------------------------------------------

\acrodef{EVS}[EVS]{Event-based Vision Sensor}
\acrodef{SNN}[SNN]{Spiking Neural Network}
\acrodef{SOTA}[SOTA]{State-Of-The-Art}
\acrodef{CNN}[CNN]{Convolutional Neural Network}
\acrodef{MLP}[MLP]{Multi-Layer Perceptron}
\acrodef{SSM}[SSM]{State Space Model}
\acrodef{RNN}[RNN]{Recurrent Neural Networks}
\acrodef{QAT}[QAT]{Quantization-Aware Training}

% -----------------------------------------------------------------------------------

\title{Privacy-preserving fall detection at the edge using Sony IMX636 event-based vision sensor and \\ Intel Loihi~2 neuromorphic processor}

\author{
    \IEEEauthorblockN{
        Lyes Khacef\IEEEauthorrefmark{1}\thanks{Corresponding author: lyes.khacef@sony.com},  % dataset characterization, training and sim+Loihi~2 eval of SNN models, retraining of S4D-based models, demo, initial paper writing, project coordination (SonyEU)
        Philipp Weidel\IEEEauthorrefmark{3},  % development of Loihi-compatible MCUNet model, training and sim+Loihi~2 eval of S4D-based models, SW delivery, demo
        Susumu Hogyoku\IEEEauthorrefmark{2},  % dataset creation, demo, project coordination (SonyJP)
        Harry Liu\IEEEauthorrefmark{5},  % HW interface development
        Claire Alexandra Br\"auer\IEEEauthorrefmark{1},  % training and sim eval of SNN models
        Shunsuke Koshino\IEEEauthorrefmark{2}, \\  % dataset anonymization and delivery
        Takeshi Oyakawa\IEEEauthorrefmark{2},  % HW characterization
        Vincent Parret\IEEEauthorrefmark{1},  % project management (SonyEU)
        Yoshitaka Miyatani\IEEEauthorrefmark{2},  % project management (SonyJP)
        Mike Davies\IEEEauthorrefmark{5},  % project management (Intel)
        Mathis Richter\IEEEauthorrefmark{4}  % HW delivery, project coordination (Intel)
    }
    \IEEEauthorblockA{
        \IEEEauthorrefmark{1}
        \itshape
        Sony Advanced Visual Sensing AG, Schlieren, Switzerland
    }
    \IEEEauthorblockA{
        \IEEEauthorrefmark{2}
        \itshape
        Sony Semiconductor Solutions Corporation, Research Division 1, Tokyo, Japan
    }
    \IEEEauthorblockA{
        \IEEEauthorrefmark{3}
        \itshape
        Intel Labs, Intel Semiconductor AG, Zurich, Switzerland
    }
    \IEEEauthorblockA{
        \IEEEauthorrefmark{4}
        \itshape
        Intel Labs, Intel Deutschland GmbH, Feldkirchen, Germany
    }
    \IEEEauthorblockA{
        \IEEEauthorrefmark{5}
        \itshape
        Intel Labs, Intel Corporation, Santa Clara, CA, USA
    }
}

\maketitle

% -----------------------------------------------------------------------------------

\begin{abstract}
Fall detection for elderly care using non-invasive vision-based systems remains an important yet unsolved problem.
Driven by strict privacy requirements, inference must run at the edge of the vision sensor, demanding robust, real-time, and always-on perception under tight hardware constraints.
To address these challenges, we propose a neuromorphic fall detection system that integrates the Sony IMX636 event-based vision sensor with the Intel Loihi~2 neuromorphic processor via a dedicated FPGA-based interface, leveraging the sparsity of event data together with near-memory asynchronous processing.
Using a newly recorded dataset under diverse environmental conditions, we explore the design space of sparse neural networks deployable on a single Loihi~2 chip and analyze the tradeoffs between detection $\boldsymbol{F_1}$ score and computational cost. 
Notably, on the Pareto front, our LIF-based convolutional SNN with graded spikes achieves the highest computational efficiency, reaching a 55$\times$ synaptic operations sparsity for an $\boldsymbol{F_1}$ score of 58\%.
The LIF with graded spikes shows a gain of 6\% in $\boldsymbol{F_1}$ score with 5$\times$ less operations compared to binary spikes.
Furthermore, our MCUNet feature extractor with patched inference, combined with the S4D state space model, achieves the highest $\boldsymbol{F_1}$ score of 84\% with a synaptic operations sparsity of 2$\times$ and a total power consumption of 90~mW on Loihi~2. 
Overall, our smart security camera proof-of-concept highlights the potential of integrating neuromorphic sensing and processing for edge AI applications where latency, energy consumption, and privacy are critical.
\end{abstract}


% -----------------------------------------------------------------------------------

\section{Introduction}
The world population is rapidly growing with an increase in life expectancy, particularly in developed countries~\cite{Bloom_Luca16}. Consequently, elderly care is a global concern which is gaining attention. 
In particular, falls are becoming a major public health problem for the elderly~\cite{Tasnim_Eisuke23}. According to the World Health Organization, falls are the second most common cause of unintentional injury-related deaths worldwide, after road traffic accidents~\cite{WHO_21}. 
Therefore, there is a growing need for fall detection systems in the healthcare industry.
A fall detection system aims to trigger an alert \textit{after} a fall has occurred to emergency caregivers~\cite{Wang_etal20}. 
On the one hand, quick response is critical after a fall and medical outcomes of falls as well as the associated healthcare expenses largely depend on rescue time~\cite{Mubashir_etal13,Gharghan_Hashim24}. 
On the other hand, it has been shown that fall detection systems decrease the fear of falling, which in turn can reduce the likelihood of a fall~\cite{Friedman_etal02,Mubashir_etal13}.

Fall detection systems use two main types of sensors~\cite{Chen_etal17}: body-worn inertial sensors and vision sensors.
Accelerometers and gyroscopes embedded in smartphones and smartwatches measure body motion independently of environmental factors (e.g., objects, lighting). However, continuous wear and regular recharging limit their practicality, particularly for elderly people~\cite{Alam_etal22}.
In contrast, vision sensors like RGB cameras, infrared cameras, and depth cameras are non-intrusive for the user and provide continuous passive monitoring without depending on user compliance. They are therefore useful in settings such as nursing homes, hospitals, or assisted living facilities. 
Combined with deep learning models, vision-based solutions can be highly accurate and adaptable to different environments~\cite{Shojaei-Hashemi_etal18}. Yet, privacy concerns remain substantial~\cite{Serpanos_etal08,Igual_etal13,Prasad_etal24} as residents often oppose in-room cameras~\cite{Xu_etal18}.
Studies on fall detection usually lack strategies to effectively ensure data privacy, which prevents their deployment in real-world settings~\cite{Igual_etal13}.

In this work, we propose a privacy-preserving vision-based solution that consists of the Sony\textregistered{}\footnote{Sony is a registered trademark of Sony Group Corporation or its affiliates.} \ac{EVS}~\cite{Finateu_etal20} which only captures per-pixel changes of light intensity in the form of events, combined with the Intel\textregistered{}\footnote{Intel is a registered trademark of Intel Corporation or its affiliates.} Loihi~2~\cite{Davies_etal21} which directly processes these events asynchronously in a fast and efficient way.
This system effectively preserves privacy by processing the sparse event stream in real time, enabling rapid detection while ensuring that events are discarded immediately after processing at the edge. 
In addition, it reduces energy consumption by exploiting the sparsity of input events and neural networks activations which is important for always-on inference, especially when a battery-powered operation is needed.
Beyond the fall detection application, this smart security camera is a proof-of-concept of real-time always-on perception systems at the edge where latency, energy consumption, and privacy matter.

In the last five years, event-based vision sensing has been extensively explored to solve many classes of computer vision problems such as classification, segmentation, detection, and tracking~\cite{Gallego_etal22,Cordone_etal22,Cazzato_Bono24,Chen_etal25,Yang_etal25}. However, when targeting a fully embedded system, \ac{EVS} requires dedicated processing to realize the latency and energy efficiency advantages of the sensor at the system level.
In particular, asynchronous near-memory processing using digital neuromorphic chips is a natural fit in terms of processing paradigm and technology maturity to exploit the high temporal resolution and sparsity of event streams \cite{Christensen_etal22,Schuman_etal22,Kudithipudi_et25}.
This end-to-end neuromorphic sensing and processing approach was particularly studied using the Intel Loihi research chip~\cite{Davies_etal21}, which was shown to be 3-30$\times$ more energy-efficient than the embedded Nvidia GPUs~\cite{Paredes-Valles_etal24, Ceolini_etal20}.

The main limitation of these works is the USB-based interface, which allows for high flexibility but imposes bottlenecks on the speed and efficiency of the system.
Therefore, there is growing interest for direct interfaces between \ac{EVS} and embedded processors.
For example, dedicated interfaces have been developed to connect \ac{EVS} with FPGA-based processing~\cite{Bonazzi_etal25}. However, these systems remain power-hungry as they target programmability over efficiency.
Other systems in research~\cite{Rutishauser_etal23} and industry~\cite{Yao_etal24,Caccavella_etal24} propose dedicated interfaces on ASICs, but they use low-resolution \ac{EVS} (typically $128 \times 128$ pixels) and embed very constrained processors, limiting the system's accuracy and robustness for complex computer vision tasks.

This work presents the following key contributions toward a robust, privacy-preserving embedded fall-detection system:
\begin{itemize}
    \item We present a new FPGA-based interface that directly connects the IMX636 sensor with the Loihi~2 processor.
    \item Using a newly recorded dataset, we explore and benchmark different algorithmic solutions on a single Loihi~2 chip, evaluating detection $F_1$ score, latency, computational sparsity/cost, and power consumption. 
    \item We introduce graded spikes for Leaky Integrate-and-Fire (LIF) models and investigate their impact in \acp{SNN}, achieving the highest sparsity. 
    \item We deploy for the first time an MCUNet architecture on neuromorphic hardware using patched inference to fit in a single Loihi~2 chip. We combine it with a State Space Model (SSM), achieving the highest $F_1$ score.
    \item We provide insights into the trade-offs along the Pareto front between $F_1$ score and computational cost, demonstrating the possibilities of integrating the IMX636 sensor with the Loihi~2 processor.
\end{itemize}

% -----------------------------------------------------------------------------------

\section{Sensing and processing technologies}
\label{sec:sensor_processor}
We introduce the IMX636 \ac{EVS} sensor and the Loihi~2 asynchronous processor, which we integrate using a direct FPGA-based interface to realize the low-latency and low-power advantages of \ac{EVS} at the system level.

\begin{center}
    \begin{figure}[!htbp]
        \includegraphics[width=\linewidth]
        {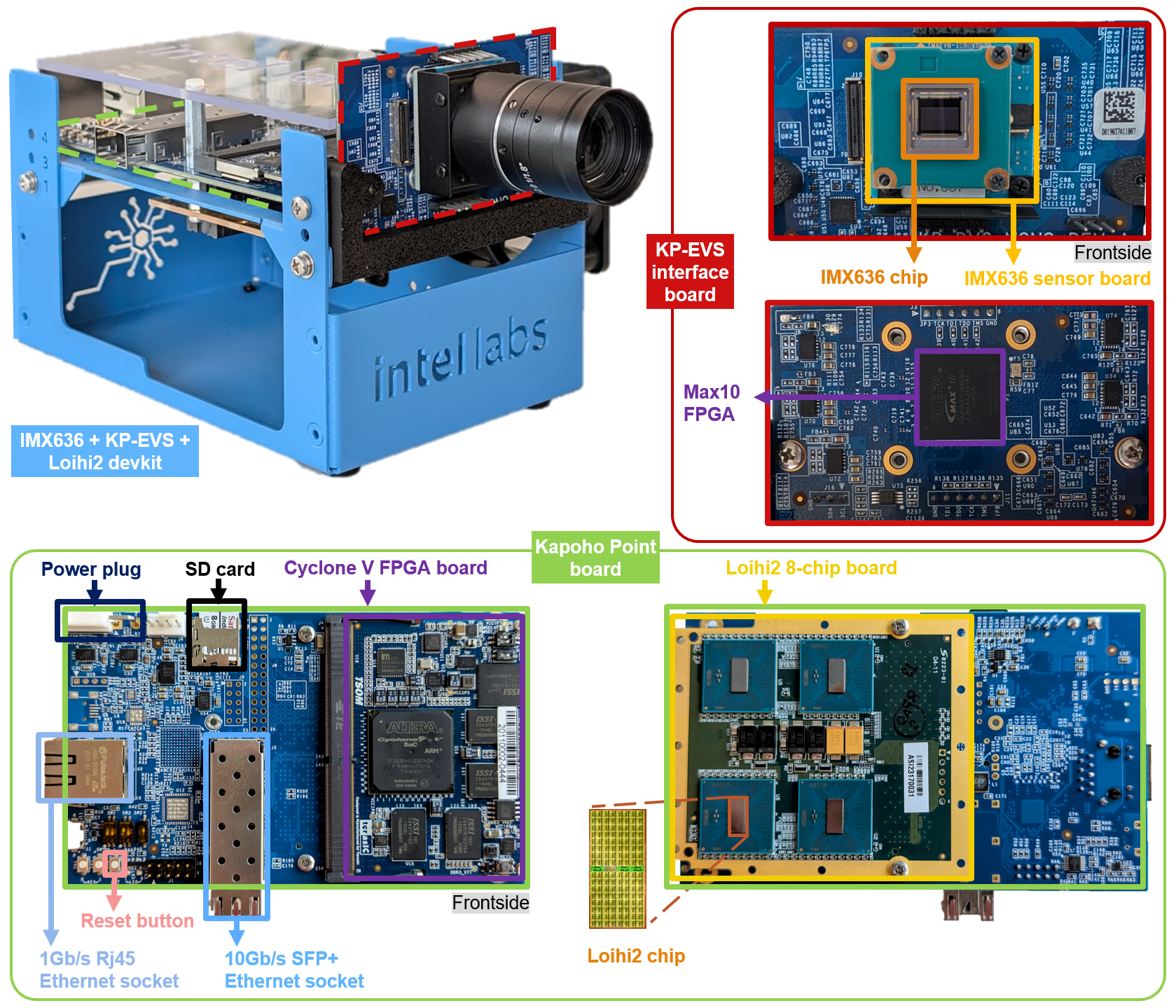}
        \centering
        \caption{Hardware system overview showing the whole system \textit{(top left)}, the KP-EVS interface board \textit{(top right)}, the host board with IO and power interfaces \textit{(bottom left)}, and a Kapoho Point (KP) board with 8 Loihi~2 chips (only one chip is used in our work) \textit{(bottom right)}.}
        \label{fig:hw_system}
    \end{figure}
\end{center}

\subsection{Sony IMX636 sensor}
\ac{EVS} are bio-inspired sensors that asynchronously output per-pixel luminance changes in the form of events. Each pixel independently emits an event when the log-brightness difference since the last event reaches a threshold~\cite{Gallego_etal22}.
An event $i$ is denoted by ${e}_i = (x_i, y_i, t_i, p_i)$, where $x_i$ and $y_i$ are the pixel coordinates, $t_i$ is the timestamp with $\mu$s resolution and $p_i \in \{+,-\}$ is the polarity, i.e., positive or negative for an increase or decrease in light intensity, respectively. 
This differential sensing paradigm leverages the redundancy in the natural environment, and outputs sparse events with low latency (less than 100~$\mu$s at 1 KLux), high temporal resolution ($\mu$s), low power (32~mW at 0.1~M events/s, 73~mW at 300~M events/s) and high dynamic range (beyond 120 dB)~\cite{Sony_EVS,Prophesee_IMX636}.

The IMX636 sensor~\cite{Finateu_etal20} was developed in a collaboration between Sony and Prophesee, combining Sony Semiconductor Solutions's CMOS image sensor technology with Prophesee's event-based vision sensing technology. 
In particular, the light-receiving and luminance detection circuits 
are incorporated on different layers: a pixel chip with a programmable Region of Interest (ROI), and a logic chip which includes integrated signal processing circuits (e.g., anti-flicker, event filter, and event rate control). These chips are stacked and connected using Cu-Cu technology within each pixel.
The IMX636 sensor has an HD resolution of 1280$\times$720 pixels, using the industry’s smallest pixel of 4.86~$\mu$m among stacked \ac{EVS}, integrated with the logic chip in a 40~nm process~\cite{Sony_EVS,Prophesee_IMX636}.

\subsection{Intel Loihi~2 processor}
Loihi~2 is the second generation of the Loihi research chip~\cite{Davies_etal18,Intel_Loihi2}, a fully digital asynchronous chip manufactured using Intel~4 process technology. 
It uses 128 near-memory asynchronous processing neuro-cores and 2 networks-on-chip for sparse communication among these cores. 
It also includes 6 embedded, synchronous x86 Lakemont cores for management, such as network configuration and data encoding/decoding. 
Multiple Loihi~2 chips can be combined to form larger systems; Fig.~\ref{fig:hw_system} shows a KP system with 8 Loihi~2 chips.

Loihi~2 uses barrier synchronization between cores to ensure that all neurons compute within the same global algorithmic timestep in a deterministic way \cite{Davies_etal18,Li_etal25}.
With stored data, the chip operates as fast as possible, whereas with streamed data, the chip can operate based on physical time (using the clock of the x86 cores), event count, or any signal from an external host or device.
In addition, algorithmically, Loihi~2 supports sparse (beyond conventional spiking) neural network deployment due to its programmable neuro-core and graded spike message-passing~\cite{Shrestha_etal24}.
The Loihi~2 inference can therefore match the accuracy of the same quantized model on an ANN accelerator, while achieving higher computing efficiency by exploiting the unstructured sparsity~\cite{Zhou_etal25,Yik_etal25_2} of synaptic operations (SynOps).
Using Loihi 2, we aim not for biological plausibility, but for energy efficiency via biological inspiration.

% -----------------------------------------------------------------------------------

\section{Methods}
\label{sec:methods}
This section describes the FPGA-based hardware interface and defines our algorithmic exploration with four neural architectures of less than a million parameters which can be deployed on a single Loihi~2 chip (CNN, MLP, MCUNet, S4D), five neuron models (ReLU, SigmaDelta, binary LIF, graded LIF, SSM), as well as the Loihi~2-aware patched inference and training methods.

\subsection{Dedicated hardware interface}
\begin{center}
    \begin{figure}[!htbp]
        \includegraphics[width=\linewidth]
        {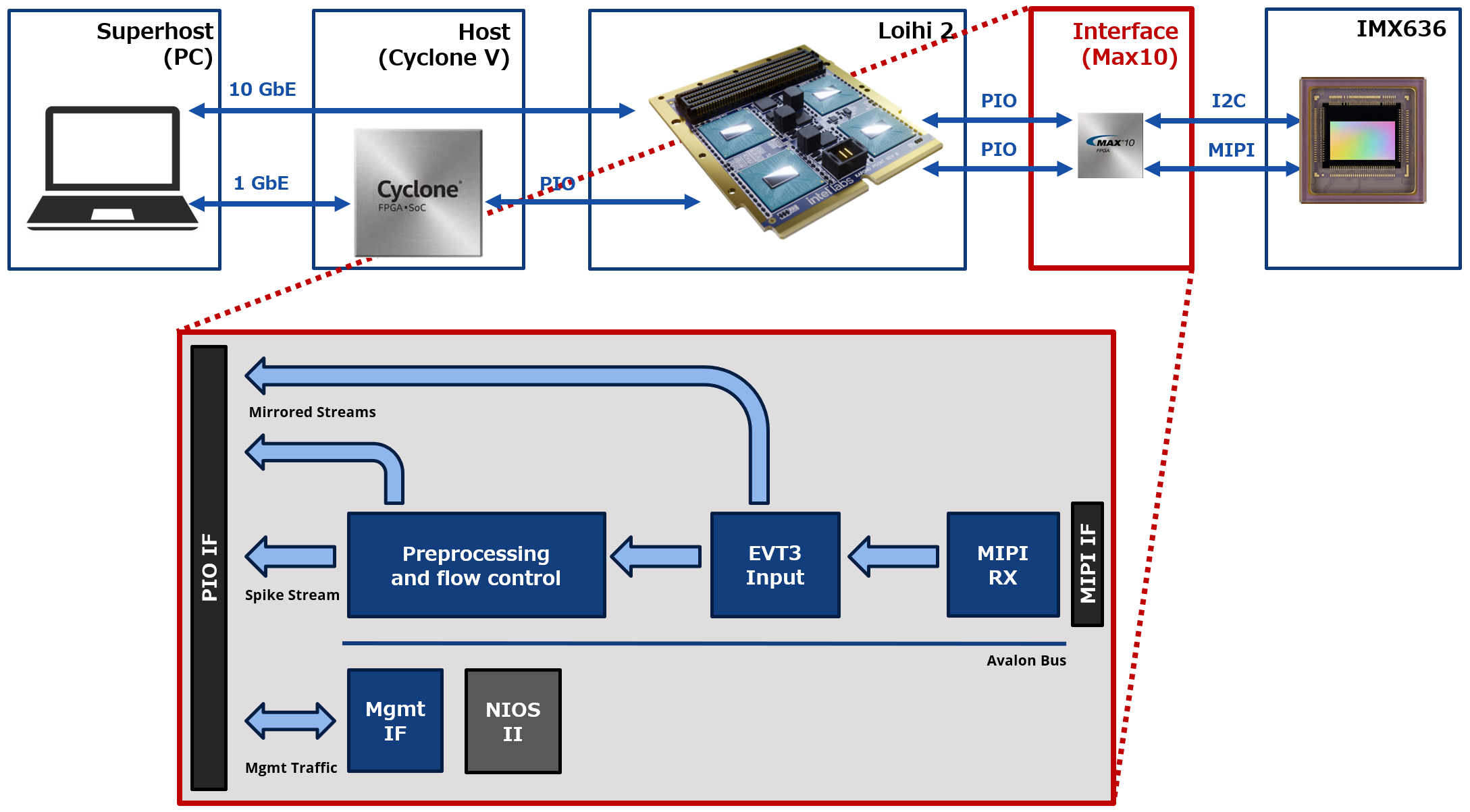}
        \centering
        \caption{Hardware system pipeline with details of Max10 FPGA interface.}
        \label{fig:fpga_interface}
    \end{figure}
\end{center}

The KP-EVS dedicated hardware interface connects the IMX636 sensor chip to the KP processing system via an interface board (top right of Fig.~\ref{fig:hw_system}, Fig.~\ref{fig:fpga_interface}). 
The logic is implemented on a Max10 10M50 FPGA from Altera. It accepts EVT3~events from the IMX636 sensor over MIPI~CSI\nobreakdash-2 (2~lanes at 600~Mbps), applies cropping (drop events outside a rectangular ROI) and down-sampling (reduce spatial resolution), maps each event to an algorithmic timestep and up to 4 neuro cores to spike to, synchronizes with Loihi~2, and forwards the spikes via Parallel Input-Output (PIO) at the required timestep. 
A Nios~II embedded processor core supports management access for configuring event processing and spike generation, and for configuring and controlling the IMX636 sensor through~I2C.
Validation and visualization is supported by an optional mirroring of received EVT3 events and generated spikes.
The full design maps to 17K~logical elements and 128~KB of memory, and runs at 100~MHz. 

\subsection{Neural network architectures}
\textbf{CNN.}
We use the following 5-layer \ac{CNN} topology: $160 \times 160 \times 2 - 16c3 - 32x3 - 64c3 - 128c3 - 256c3$, where $XcY$ denotes a convolution layer with $X$ kernels of shape $Y \times Y$. All convolution layers have a stride of $2$.
The output features have a size of $4 \times 4 \times 256$.

\textbf{MLP.}
On top of the \ac{CNN} and after a flattening layer, a 3-layer \ac{MLP} is used with the following fully-connected topology: $128 - 64 - 2$. Two output neurons are used to represent the classes `Fall' and `NoFall'.

\textbf{MCUNet.}
MCUNet is a \ac{CNN}-based model designed to run on tiny hardware resources for always-on inference in IoT devices~\cite{Lin_etal20,Lin_etal21}.
It has 18~MobileNet inverted residual blocks~\cite{Sandler_etal18}, characterized by depthwise separable convolution~\cite{Brito_etal25}, linear bottlenecks, inverted residuals, and non-linear activation.
Nevertheless, it is possible to use an arbitrary number of blocks, which is denoted by MCUXB, where X refers to the number of blocks.
For instance, we only use 13 blocks to deploy the model on a single Loihi~2 chip.
Notably, this is the first time that such a \ac{SOTA} CNN architecture is deployed on neuromorphic hardware.

\textbf{S4D.}
\label{sec:s4d_network}
On top of the stateless \ac{CNN}/MCUNet spatial feature extractors and after average pooling and flattening, a \ac{SSM} is used for temporal feature extraction.
\acp{SSM} have recently been positioned as a promising alternative to Transformers for sequence modeling~\cite{Patro_Agneeswaran25}.
In their convolutional form, unlike recurrent neural networks, they enable highly parallel training on GPUs with stored data.
In their recurrent form, they do not suffer from the quadratic scaling of compute cost of the attention mechanism of Transformers, allowing for linear scaling and efficient inference with streaming data.
The S4D model~\cite{Gu_etal22} is a variant of the Structured State Space for Sequence Modeling (S4) architecture, where a diagonal state space matrix is used to simplify the kernel computation. It was successfully deployed on Loihi~2~\cite{Meyer_etal24}.

\subsection{Neuron models}
For all neuron models, let us consider the activation $z$ at the algorithmic time step $t$ as follows:
\begin{equation*}
    z[t] = \sum_{i=1}^{n} w_i \times x_i[t] + b
\end{equation*}
where $w_i$ and $x_i$ are the weight and the input of the synapse $i$, respectively, $n$ is the number of synapses, and $b$ is a bias.

Let us also define spiking neurons as stateful neurons with sparse activations in the form of binary or graded (i.e., valued) spikes, including thus SigmaDelta and LIF neurons.

\textbf{ReLU.}
ReLU neurons output $y[t]$ is defined as follows:
\begin{equation*}
    y[t] = 
    \begin{cases}
        z[t] & \text{if } z[t] \geq 0 \\
        0 & \text{otherwise}
    \end{cases}
\end{equation*}
It adds non-linearity by zeroing the negative activations, which introduces some inherent sparsity.

\textbf{SigmaDelta.}
SigmaDelta neurons aim to further sparsify the ReLU output with the stateful Delta encoder as follows:
\begin{align*}
    a[t] &= 
    \begin{cases}
        z[t] & \text{if } z[t] \geq 0 \\
        0 & \text{otherwise}
    \end{cases} \\
    \Delta a[t] &= a[t] - a[t-1] + r[t-1] \\
    y[t] &=
    \begin{cases}
        \Delta a[t] & \text{if } \Delta a[t] \geq \vartheta \\
        0 & \text{otherwise}
    \end{cases} \\
    r[t] &= \Delta a[t] - y[t]
\end{align*}
where a[t] is the ReLU output, $r[t]$ is the residual state, $y[t]$ is the output and $\vartheta$ is the learnable spike threshold.
The Delta encoder sends a graded spike with a value of $y[t]$ when $\Delta a[t]$ exceeds the threshold $\vartheta$, while the Sigma decoder receives the sparse event messages and accumulates them to restore the original information.

\textbf{LIF with binary/graded spikes.}
LIF neurons are behavioral models of biological neurons, derived from the spike response model~\cite{Gerstner_Kistler02}. 
We use the current-based LIF neuron which is useful for learning patterns with a rich temporal structure~\cite{Bouanane_etal23}. The LIF dynamics are expressed as follows:
\begin{align*}
    i[t] &= \alpha \times i[t-1] + z[t] \\
    u[t] &= \beta \times u[t-1] \times (1-\mathcal{H}(u[t-1] - \vartheta)) + i[t] \\
    y[t] &= 
    \begin{cases}
        u[t] & \text{if } u[t] \geq \vartheta \text{ and \textit{graded spikes}} \\
        1 & \text{if } u[t] \geq \vartheta \text{ and \textit{binary spikes}} \\
        0 & \text{otherwise}
    \end{cases}
\end{align*}
where $i[t]$ and $u[t]$ are the current and voltage states respectively, $\alpha$ and $\beta$ are the current and voltage decays respectively, $\mathcal{H}$ is the Heaviside step function used for the hard reset, $\vartheta$ is the learnable spike threshold, and $y[t]$ is the output spike which can be binary or graded depending on the LIF model.

\textbf{SSM.}
We implement the SSM neuron model in two different forms, as explained in Sec. \ref{sec:s4d_network}.
For training, we use the convolutional form of the S4D model allowing parallel and much faster execution on GPUs.
% when $\bm{T}$ datapoints are available, as follows: 
% \begin{align*}
%     \bm{K} &= (\bm{CB}, \bm{C A B}, \cdots, \bm{C}\bm{A}^{T-1}\bm{B}) \\
%     (\cdots, y[t], \cdots, y[T]) &= \bm{K} \ast (\cdots, z[t], \cdots, z[T]) \label{eq:s4_convolution} \\
% \end{align*}  % Lyes: I changed s[t] to z[T])
For inference, we represent the S4D dynamics in its time-discrete recurrent formalization as follows: 
\begin{align*}
    s[t] &= a \times s[t-1] + b \times z[t] \\
    y[t] &= c \times s[t]
\end{align*}

where $s[t]$ is the neuron's state, $a$, $b$ and $c$ are the local parameters and $y[t]$ is the output.
In fact, the recurrent weight matrix $A$, the input projection matrix $B$, and the output projection matrix $C$ of the S4D model are diagonalized, so their diagonal values become local dynamical parameters per neuron, analogous to the synaptic and membrane decay factors (or time constants) of LIF neurons~\cite{Bal_Sengupta25,Fabre_etal25}.
% The major difference to the LIF neuron is that the parameters in our S4D neuron are complex numbers, which results in oscillatory behavior.

\subsection{Patched inference}

\begin{center}
    \begin{figure}[!htbp]
        \includegraphics[width=\linewidth]
        {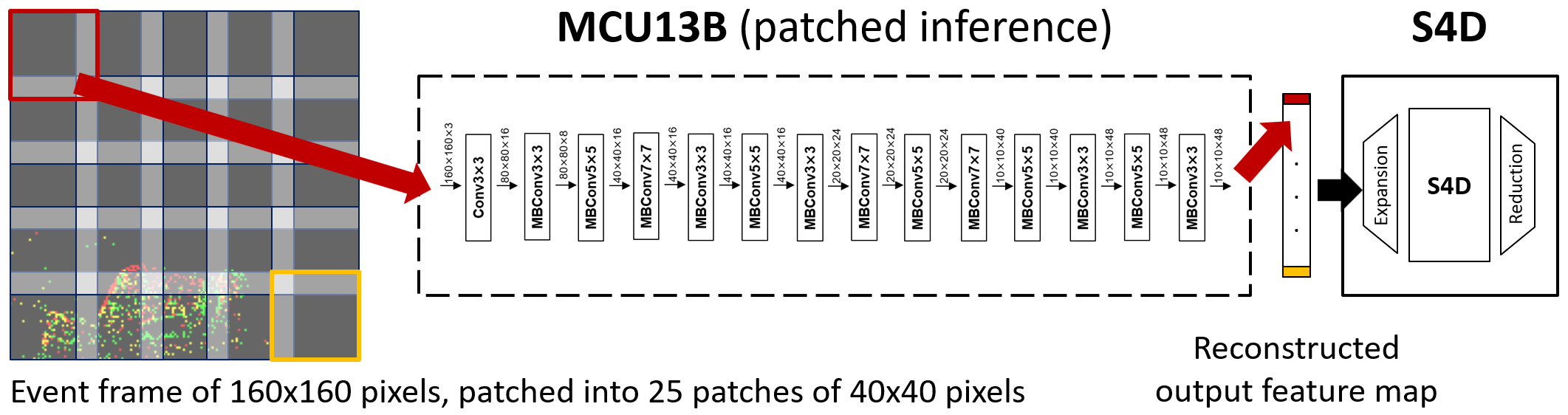}
        \centering
        \caption{Input-patched inference of MCU13B model for a single event frame, which reduces the memory requirements by an order of magnitude by reusing the Loihi~2 neuro-core memories for each patch.}
        \label{fig:mcu_patching}
    \end{figure}
\end{center}

Despite using only 13 blocks of MCUNet, we need around 10 Loihi~2 chips to deploy the entire network due to the large feature map sizes (i.e., number of neurons).
However, since the model is stateless, there is no need to keep the neurons states for consecutive time steps.
Therefore, we adapt and use the patched inference proposed in~\cite{Lin_etal21}, where the authors split the input into multiple \textit{patches}, and process multiple layers per patch to reduce the peak memory requirement of the model.
Similarly, we split the 160x160 pixels input into 25 patches of 40x40 pixels as shown in Fig.~\ref{fig:mcu_patching}, with a stride of 30 pixels (i.e., overlap of 10 pixels).

The patches are processed sequentially, where each patch goes through the whole 13 blocks of MCU13B (not just a few layers as in~\cite{Lin_etal21}) and only the final output feature map is saved.
By reusing the Loihi~2 neuro-core memories for each patch, this input-patched inference drastically reduces the memory requirements by an order of magnitude.
Once all the patches are processed, the complete output feature map is reconstructed and the S4D inference is triggered.
It is to note that the 10 pixels overlap is not enough to produce algorithmically equivalent feature extraction, and this design choice was motivated by the need to fit the model in a single Loihi~2 chip. It is therefore expected to have some loss in feature map quality and overall detection accuracy.

\subsection{Training}
\textbf{Events encoding.}
The asynchronous events produced by \ac{EVS} must be \textit{accumulated} into algorithmic timesteps for inference on Loihi~2 and for training on GPUs.
We accumulate events into constant temporal windows (i.e., algorithmic timestep duration) and apply a per-pixel and per-polarity count to get a single graded input spike. The only exception is the binary LIF model where we use binary input spikes (i.e., at least one event was received).
During Loihi~2 timestep-based inference, event frames are not built, and events are simply encoded into sparse graded spikes.

We design the fall detection system to operate at a fixed throughput of 16~Hz.
Accordingly, we use the maximum timestep duration of 60~ms for the S4D-based models to accommodate the processing latency of the patched MCUNet backbone.
For the spike-based models, we use a shorter timestep duration of 20~ms to enable finer-grained temporal processing, while accumulating spikes over three consecutive timesteps to maintain the same detection throughput.

\textbf{Focal loss.}
Our fall detection dataset is highly imbalanced ($7\%$ of falling actions); therefore, we adopt the Focal Loss (FL)~\cite{Lin_etal20_loss} to mitigate class imbalance, defined as follows:
\begin{equation*}
    \text{FL}(p,y) = 
    \begin{cases}
        -\alpha (1-p)^{\gamma} \log(p), & \text{if } \hat{y} = 1, \\[6pt]
        -(1-\alpha) p^{\gamma} \log(1-p), & \text{otherwise}.
    \end{cases}
\end{equation*}
where $p \in [0,1]$ is the model's estimated probability for the fall class with label $\hat{y}=1$, $\hat{y} \in \{0,1\}$ is the ground-truth class label, $\alpha \in [0,1]$ is the balancing factor to address class imbalance, and $\gamma \geq 0$ is the focusing parameter that reduces the relative loss for well-classified samples.

The model's estimated probability $p$ is calculated over the sample's duration based on the sum of output spikes for the spike-based models, whereas it is based on the maximum logit difference the continuous-output models.

\textbf{Backpropagation.}
While ReLU- and S4D-based models are trained using standard Backpropagation (BP), LIF and SigmaDelta spiking models require a custom BP to address the non-differentiability limitation of their activation functions \cite{Neftci_etal19,Eshraghian_etal23}.
In this work, we use the SLAYER framework~\cite{Shrestha_Orchard18} for direct training of \acp{SNN}, as it is widely used and takes into account the hardware constraints (e.g., weight and state quantization) of the Loihi~2 chip.

\begin{center}
    \begin{figure}[!htbp]
        \includegraphics[width=\linewidth]
        {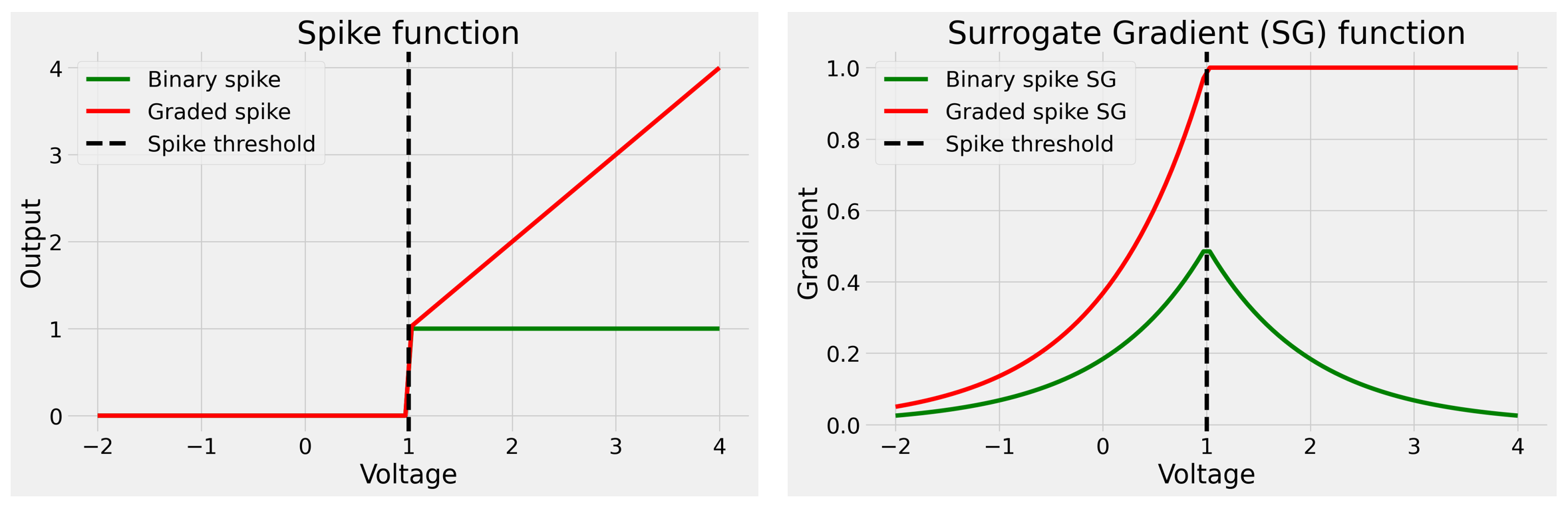}
        \centering
        \caption{Spike (activation) functions \textit{(left)} and surrogate gradient functions \textit{(right)} of LIF neurons with binary and graded spikes.}
        \label{fig:surrogate_grads}
    \end{figure}
\end{center}

SLAYER is a form of Backpropagation Through Time (BPTT) which uses surrogate gradients in the backward pass~\cite{Neftci_etal19}, as shown in Fig.~\ref{fig:surrogate_grads}.
In addition, gradients are computed jointly in time, reducing the complexity of sequential BPTT by parallelized and GPU-accelerated joint gradient computation~\cite{Bauer_etal23}. Although it is not mathematically equivalent to BPTT, SLAYER does not differ significantly, in particular when the spiking rate is low.

\textbf{Hyper-parameters.}
All models were trained for 100 epochs with a batch size of 16, using $\alpha$ = 0.9 and $\gamma$ = 2.0 for the focal loss fuction. 
The CNN+MLP models were trained with \ac{QAT} using 8-bit weights and a learning rate of 1e-5. 
The CNN+S4D and MCUNet+S4D were trained with full precision using the convolutional representation of S4D, then quantized and fine-tuned for 10 epochs with \ac{QAT} using 8-bit weights, a backbone learning rate of 5e-5 and a S4D learning rate of 5e-6.

% -----------------------------------------------------------------------------------

\section{Experiments and results}
\label{sec:results}
In this section, we present our Loihi~2-aware algorithmic exploration and benchmarking with our newly recorded dataset, highlighting the tradeoffs between accuracy and efficiency.

\subsection{Fall detection benchmark}
\textbf{Original dataset.}
In order to have full control over the environmental conditions, we recorded a new dataset for fall detection, comprising 14 action classes \textit{(falling, seated, standing, sit down, stand up, lying, pick up, touching head, touching back, touching torso, touching neck, coughing, walking, no person)} with about 640 samples (sequences) of 4.6~s per class. 
In our experiments, we only consider the \textit{falling/no-falling} (imbalanced) binary classification problem.

% Dataset recording was outsourced to an external company to ensure quality and safety.
Dataset recording was performed in 5 locations with different backgrounds (black, white, living room, sun light, water from fountain), light conditions (source, brightness from 10 Lux to 300 Lux), and distance to the camera.
Each sample is about 4.6~s long, however we only use the 2~s with most events per sample for training and validation, while we use the full samples for testing.
The dataset was split for cross-subject evaluation, with 3906, 3182, 1793 samples for training, validation and test sets, respectively.

In order to deploy the models in a single Loihi~2 chip, we center-cropped the original 1280x720 resolution to 640x640 then downsampled it with 2x2 sum pooling to 160x160. 
In hardware, the cropping is implemented by defining an ROI on the IMX636 sensor directly, while the downsampling is implemented in the interface board with a simple integer (floor) division by 4 of the $x$ and $y$ coordinates (i.e., keeping the same events throughput).
Across the dataset, the class \textit{walking} has the most input events, with an average throughput of 1 million events/s and a peak throughput of 9 million events/s.
It is important to note that events timestamps are not used in the real-world inference, however they are still needed to store the dataset and use it in simulation.

\textbf{Metrics.}
We follow the benchmarking principles defined in NeuroBench~\cite{Yik_etal25}, with algorithmic- and system-level evaluations.
At the algorithmic level, for a pre-defined throughput of 16 predictions/s, we evaluate in software:
\begin{itemize}
    \item Detection performance: We evaluate the accuracy, precision, recall and $F_1$ score. The $F_1$ score is the most insightful metric due to the fall/no-fall classes imbalance.
    % The $F_1$ score is the most insightful metric, since it measures how well a model balances maximizing true positives while minimizing both false positives and false negatives, which is especially useful for imbalanced datasets. 
    % It is defined as follows:
    % \begin{equation*}
    %     F_1 = 2 \times \frac{\text{precision} \times \text{recall}}{\text{precision} + \text{recall}} = \frac{2TP}{2TP + FP + FN}
    % \end{equation*}
    % In our application, if a fall is not detected, it presents a threat on the person's health and leads to the loss of confidence in the system which may reduce the benefits of the detector on the fear of falling. At the same time, if the system trigger too many false positives, it can lead to device rejection by caregivers~\cite{Igual_etal13}.
    \item Synaptic operations per second (SynOps/s): We evaluate the number of SynOps/s as a measure of computational cost, and we calculate the SynOps/s sparsity which is the gain (i.e., reduction) compared to an iso-topology neural network with dense processing.
\end{itemize}

At the system-level using Loihi~2, the $F_1$ score is the same as in the algorithmic-level evaluation thanks to the Loihi~2-aware modeling and the fully digital and deterministic processing of Loihi~2. Therefore, we evaluate in hardware:
\begin{itemize}
    \item Latency: We evaluate the input to output latency, i.e., the delay between a given input timestep and the corresponding output after a full neural network inference.
    \item Number of Loihi~2 cores: It is a measure of the required silicon area, and cannot exceed 128 cores for one chip.
    \item Power consumption: We measure the static power and the average dynamic power over the test set samples.
\end{itemize}

\subsection{Algorithmic-level evaluation}
\label{sec:algo_eval}
The Loihi~2-aware modeling and 8-bit quantization-aware training is performed using LavaDL and NxTorchEV libraries. The results are presented in Tab. \ref{tab:algo_benchmark}.

\begin{table*}
\centering
    \caption{Algorithmic-level evaluation results.} 
    \label{tab:algo_benchmark}
    \resizebox{\textwidth}{!}{
    \begin{tabular}{|ll|cccc|cc|}
        \hline
        \multirow{2}{*}{Network architecture} & \multirow{2}{*}{Neuron model} & \multicolumn{4}{c|}{Test performance} & \multicolumn{2}{c|}{SynOps/s} \\
         &  & Accuracy (\%) ↑ & Precision (\%) ↑ & Recall (\%) ↑ & $F_1$ score (\%) ↑ & Cost (M/s) ↓ & Sparsity (x) ↑ \\ \hline
        \multirow{4}{*}{CNN + MLP} & ReLU & 86.8 ± 0.2 & 32.4 ± 0.5 & \textbf{81.0 ± 1.4} & 46.3 ± 0.7 & 433 & 3.3 \\ \cline{2-8} 
         & SigmaDelta & 88.5 ± 0.4 & 33.9 ± 1.4 & 66.4 ± 3.2 & 44.9 ± 2.0 & 100 & 14.5 \\ \cline{2-8} 
         & Binary LIF & 90.8 ± 0.3 & 41.0 ± 1.1 & 70.4 ± 1.8 & 51.9 ± 1.4 & 125  & 11.6 \\ \cline{2-8} 
         & Graded LIF & 94.4 ± 0.1 & 61.6 ± 1.3 & 54.6 ± 1.0 & 58.1 ± 1.2 & \textbf{26} & \textbf{55.5} \\ \hline
        CNN + S4D & ReLU + SSM & 96.9 ± 0.1 & 81.8 ± 1.0 & 72.5 ± 0.9 & 76.9 ± 0.9 & 198 & 2.9 \\ \hline
        MCU13B + S4D & ReLU + SSM & \textbf{97.8 ± 0.1} & \textbf{87.1 ± 0.1} & 80.4 ± 0.5 & \textbf{83.6 ± 0.3} & 1059 & 2 \\ \hline
    \end{tabular}
    }
\end{table*}

\begin{figure*}[htbp]
    \centering
    \begin{minipage}[b]{0.62\textwidth}
        \centering
        \includegraphics[width=\textwidth]{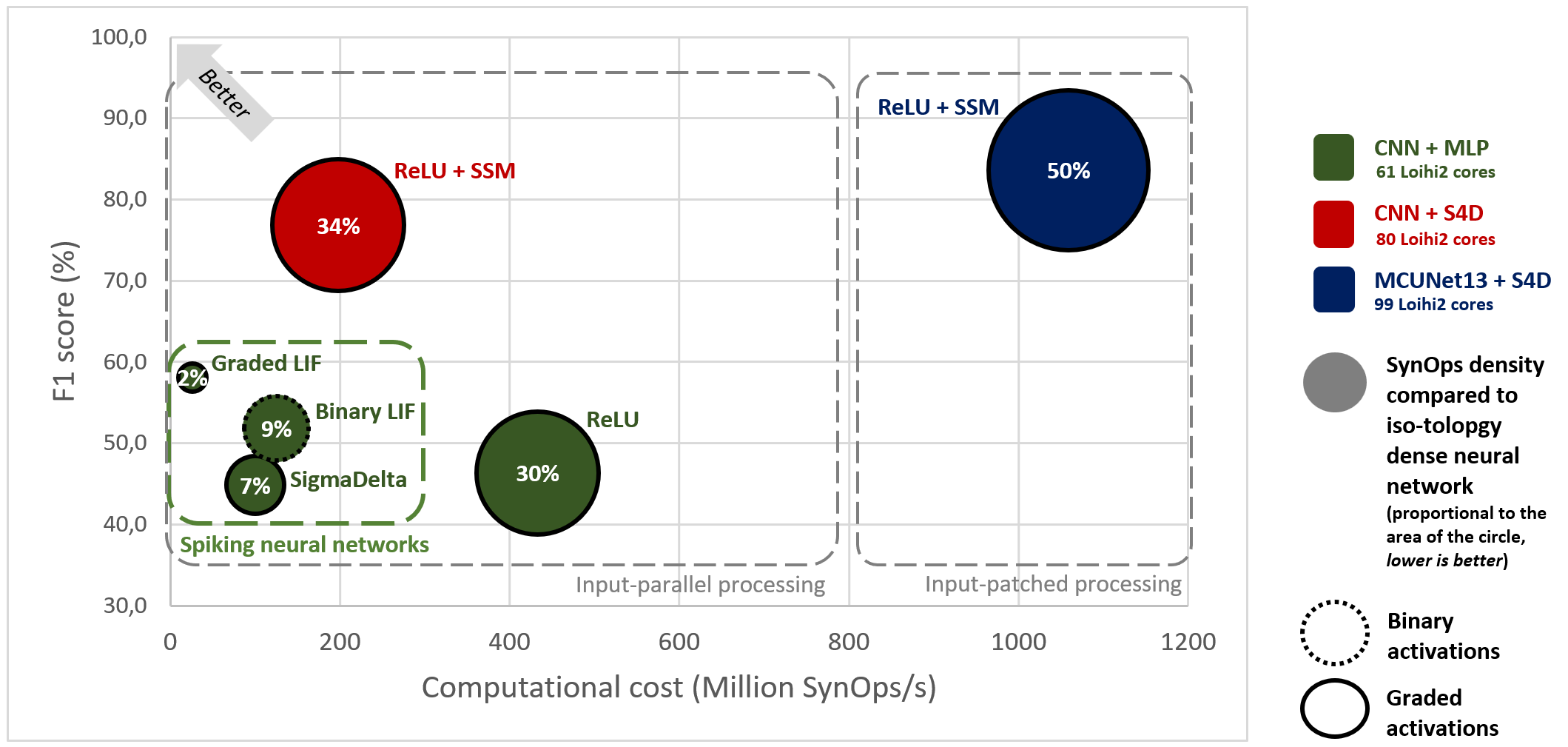}
        \caption{Fall detection algorithmic-level benchmarking at 16 predictions/s of single Loihi~2 \\ chip compatible models, highlighting the tradeoffs between $F_1$ score and computational cost.}
        \label{fig:algo_benchmark}
    \end{minipage}
    \begin{minipage}[b]{0.37\textwidth}
        \centering
        \includegraphics[width=\textwidth]{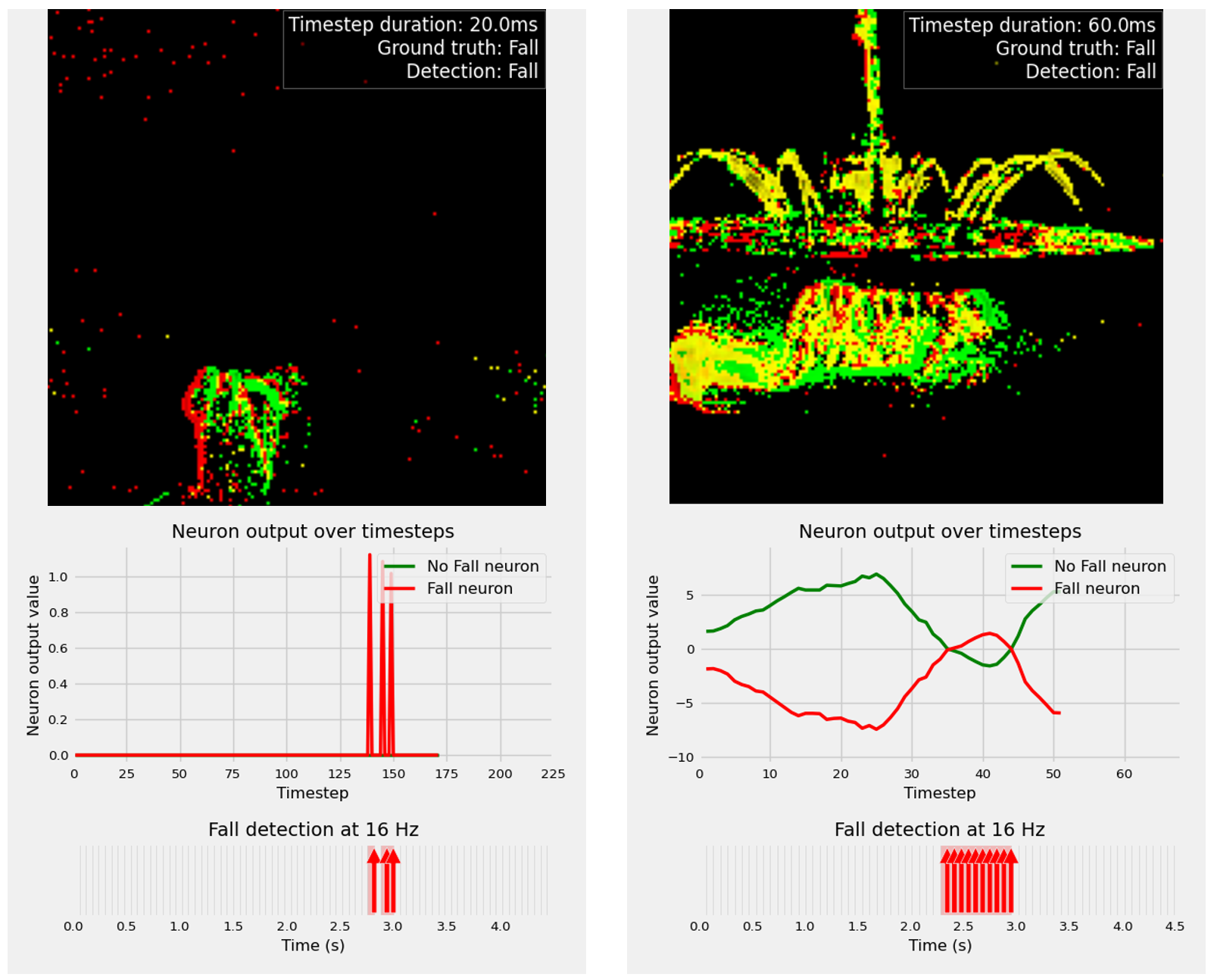}
        \caption{Events at 3.4s, logits and detection of CNN+MLP with graded LIF \textit{(left)} and MCU13B+S4D \textit{(right)}.}
        \label{fig:falls_viz}
    \end{minipage}
\end{figure*}

Within the CNN+MLP architecture, the ReLU model has the lowest $F_1$ score of 46.3\% due to the stateless processing where only individual event frames are perceived (which can lead to a high recall but very low precision), at the highest cost of 433~M SynOps/s. 
Using the SigmaDelta model, the $F_1$ score slightly degrades by 1.4\% because even though the model has states, they are used \textit{after} the original ReLU activation to sparsify the activations only without temporal feature extraction, which leads to a SynOps/s decrease of 4.3$\times$.
Replacing the SigmaDelta by the binary LIF model, the $F_1$ score increases by 7\% thanks to the temporal feature extraction. Computationally, the SynOps/s also increases by 25\%, however each SynOp is based on accumulation-only of synaptic weights (i.e., no multiplication) thanks to the binary spike paradigm.
Finally, moving from the binary to the graded LIF model, the $F_1$ score increases by another 6.2\% to reach 58.1\%, and the SynOps/s decreases by 4.8$\times$ to 26~M SynOps/s, reaching an overall SynOps/s sparsity of 55.5$\times$ compared to an iso-topology model with dense processing.

Next, we evaluate the impact of neural network architectures by first using the CNN with ReLU model and replacing the MLP by an S4D layer, decoupling spatial from temporal feature extraction. The $F_1$ score significantly increases by 18.8\% thanks to the temporal feature extraction capabilities of S4D, especially for long-range temporal dependencies, at a cost of 198~M SynOps/s.
Second, we replace the CNN by the MCU13B, which improves the $F_1$ score by another 6.7\% thanks to the enhanced spatial feature extraction, with a cost increase of 5.3$\times$ reaching 1059~M SynOps/s. Notably, the patched inference degrades the $F_1$ score by about 2\%, but allows the model to fit in a single Loihi~2 chip.
Overall, as shown in Fig. \ref{fig:algo_benchmark}, the two S4D-based models are in the Pareto-front of $F_1$ score and computational cost, where the MCU13B+S4D is the most accurate, in addition to the graded LIF-based CNN+MLP which is the most compute-efficient.

Fig. \ref{fig:falls_viz} shows two samples of fall sequences with sunset \textit{(left)} and fountain \textit{(right)} backgrounds, where the falls are correctly detected \textit{when} they happen using two models.
Our qualitative analysis through the test set visualization shows that the NoFall neuron of the CNN+MLP with graded LIFs mainly fires for sequences with background events or actions with a similarity to falling such as \textit{sit down}, as a way to reduce the probability of false positives. 
Even though its $F_1$ score is only 58.1\%, the model is able to correctly predict many falls across different light conditions and distance to the camera, as well as when the person is partially occluded.
The MCU13B+S4D clearly outperforms it in recall under extreme low light, especially when the person is far and/or highly occluded.
The precision of both models is degraded to varying extents due to false positives caused by actions with similar movements to falling, such as \textit{pickup}, \textit{sit down}, and \textit{lying}.

\subsection{System-level evaluation}
\label{sec:sys_eval}
The Loihi~2 deployment and inference was performed on a KP revision C2 board with a N3C1 Loihi2 chip, using the NxKernel~0.2.1 library. 
We mapped every model trying to minimize the number of Loihi~2 cores, and we verified that the KP-EVS hardware is electrically functional with an event stream passing through the interface into Loihi~2.
However, it was not used in the system-level evaluation which is based on stored data, sending recorded events from the superhost PC to the KP board using the 10G Ethernet interface.

\textbf{Latency.}
First, we need to define the two processing schemes of Loihi~2~\cite{Meyer_etal24}: (1) \textit{pipelined} processing where each layer is processed once at each hardware timestep, and (2) \textit{fall-through} processing where the input goes through all the layers first in multiple hardware timesteps before moving to the next input.
For the CNN-based models with input-parallel processing, a fall-through processing is used and a maximum throughput of 500~Hz can be achieved, with a minimum input-output latency of 2~ms.
For the MCUN13B+S4D model with input-patched processing, a pipelined processing of the MCU13B backbone is used with 38 hardware steps (13 blocks + 25 input patches), followed by a fall-through processing of the S4D blocks and output linear layers. A maximum throughput of 25~Hz can be achieved with a minimum input-output latency of 40~ms.
Therefore, all models are deployed in a single Loihi~2 chip and satisfy their real-time constraints by effectively reaching (and exceeding) the target 16 predictions/s. 
In the real-world inference, the Loihi~2 chip is slowed-down to operate in real-time, with an input-output latency of 60~ms which is negligible for the fall detection application that does not require a lower latency response.

\textbf{Power consumption.}
Tab. \ref{tab:sys_benchmark} summarizes the Loihi~2 cores usage and power consumption measurements of the three Pareto-front models from our algorithmic-level evaluation.
The static power is roughly proportional to the number of used cores, with about 0.75-0.95~mW/core depending on the used resources. Dynamic power is about 8-12 pW/(SynOp/s) (when normalized with the SynOps/s cost from Tab.~\ref{tab:algo_benchmark}), showing that dynamic energy consumption is indeed proportional to the number of SynOps thanks to the asynchronous processing of Loihi~2.
Nevertheless, static power is dominant, which reduces the computational efficiency impact of the CNN+MLP with graded LIFs (46.3~mW for 26~M SynOps/s) compared to the MCU13B+S4D (88.9~mW for 1059~M SynOps/s).

\begin{table}
\centering
    \caption{System-level benchmarking results of the algorithmic-level Pareto-front solutions.} 
    \label{tab:sys_benchmark}
    \resizebox{\linewidth}{!}{
    \begin{tabular}{|l|c|ccc|}
        \hline
        \multirow{2}{*}{Pareto front model} & \multirow{2}{*}{\begin{tabular}{c} Loihi~2 \\ cores \end{tabular}} & \multicolumn{3}{c|}{Power (mW)} \\
         &  & Static & Dynamic & Total \\ \hline
        CNN + MLP [Graded LIF] & 61 & 46.0 & 0.3 & 46.3 \\ \hline
        CNN + S4D & 80 & 76.0 & 1.5 & 77.5 \\ \hline
        MCU13B + S4D & 99 & 80.7 & 8.2 & 88.9 \\ \hline
    \end{tabular}%
    }
\end{table}

% -----------------------------------------------------------------------------------

\section{Discussion}
\label{sec:discussion}

\subsection{Improving fall detection $F_1$ score}
Sec.~\ref{sec:algo_eval} shows that the biggest improvements in $F_1$ score come from the neural network architectural changes, with the best $F_1$ score using MCU13B+S4D which decouples stateless spatial and stateful temporal feature extraction.
Notably, this is the first time that such a \ac{SOTA} architecture is deployed on neuromorphic hardware, thanks to the neuro-cores programmability and graded spike message-passing of Loihi~2.

To explore the potential of the MCUNet feature extractor, we train the full MCU18B+S4D without patching (i.e., cannot de deployed in a single Loihi~2 chip) and reach an $F_1$ score of 91\%, which is a significant improvement of 7.4\% with the additional 5 blocks.
In order to deploy the model in a single chip, we first need to adapt the MCU18B architecture because of the feature map size becoming smaller than the kernel size after the 13th block when using the current patched inference. 
Then, more resources per core and/or more cores per chip may be needed to embed the full model in a single Loihi~2 chip.

While further algorithmic improvements are likely necessary, a promising approach involves learning the variance of predictions in a self-supervised way. This enables the modeling of input-dependent aleatoric uncertainty \cite{Kendall_Yarin17}, which can encourage more robust and reliable predictions \cite{paredes-valles_etal25}.

\subsection{Reducing power consumption}
First, at the hardware level, the Loihi~2 power consumption is dominated by static power which is due to leakage current, as described in Sec. \ref{sec:sys_eval}.
In order to reduce it, a product version of the current Loihi~2 research chip could remove the unused components for a dedicated inference chip, and be manufactured using a more advanced technology process.
Once static power is reduced, dynamic power will become more impactful and reducing it will require a higher sparsity of SynOps/s.
Loihi~2 can exploit three levels of unstructured sparsity: (1) input event sparsity, which is inherent in \ac{EVS} and can be improved by adapting the sensor biases, (2) neural network connection sparsity by pruning low-weight/impact synapses and dead neurons which was not explored in this work, and (3) neural activation sparsity which we focused on in this work, however without explicit regularization.

Second, at the algorithmic level, we have seen that activation sparsity is the highest for \textit{stateful} spiking neurons, where the state is either used specifically to sparsify activations as in SigmaDelta neurons, or to further add temporal feature extraction as in LIF neurons. 
Stateful models are the most promising for achieving the highest sparsity while maintaining a high accuracy because they can exploit the context encoded in the states \cite{Zhou_etal25}, at the cost of a higher memory footprint.
While stateful processing cannot be applied to the patched MCU13B since Loihi~2 memories are overwritten for each patch, it can be applied to the CNN+S4D model where a variant of the SigmaDelta neuron can be used to sparsify the SSM neurons activations.

Third and finally, the current system is a proof-of-concept with an FPGA-based interface that can be integrated in the next Loihi silicon for a smaller and more efficient two-chip solution.
Looking forward, integrating the IMX636 sensor and Loihi~2 processor into a one-chip solution would reduce system footprint while enabling ultra-low latency feedback to the sensor, opening possibilities for scene-specific adaptive sensor control and power optimization.
However, such on-sensor integration imposes stringent memory constraints, likely requiring new designs to keep the total on-chip memory within a few hundred kilobytes for small sensors and potentially under one megabyte for larger sensors. 
These trade-offs will be critical considerations in evolving toward fully integrated neuromorphic vision systems.

\subsection{On-chip anonymization of EVS data}
Fully embedded processing is necessary for privacy-preserving fall detection, but it may not be sufficient at the application level as some \ac{EVS} data may have to be be sent to a cloud service in the case of a fall detection, for example, to confirm the need for medical assistance.
However, raw events should not be transmitted, because even without full RGB content they can be used to reconstruct grayscale frames, which undermines privacy protection.~\cite{Rebecq_etal21,Paredes-Valles_etal21,Zhu_etal22}.
Therefore, on-chip anonymization of the event data that would be transferred (e.g., a sliding-window buffer of a few seconds) is required in an always-on fashion, in parallel to the fall detection inference.

For example, authors in~\cite{Ahmad_etal23} try to anonymize event streams to protect the identity of human subjects against image reconstruction, by scrambling events and enforcing the degradation of such images.
However, deploying a similar algorithm in any embedded device remains challenging in terms of required computing resources and power.
Further research is required to get a fully privacy-preserving ecosystem which will enable the adoption of security cameras in sensitive environments, using neuromorphic sensing and processing in particular.

% -----------------------------------------------------------------------------------

\section{Conclusion}
\label{sec:conclusion}
We presented in this paper a fully embedded system that integrates the Sony IMX636 \ac{EVS} sensor with the Loihi~2 neuromorphic processor via a dedicated FPGA-based interface.
We extensively explored the design space of compatible neural network architectures and neuron models with a thorough benchmarking to understand the tradeoffs between $F_1$ score and computational cost/energy consumption, and highlighted the most promising models in the Pareto front.

We showed that the proposed graded spikes in LIF models increase the $F_1$ score by 6.2\% and decrease the number of SynOps/s by 4.8$\times$ compared to binary spikes, reaching an overall SynOps/s sparsity of 55.5$\times$ compared to an iso-topology neural network with dense processing. 
Compared to the binary LIF, the graded LIF comes at the cost of an extra multiplication for each SynOp which is nevertheless natively supported by Loihi~2 thanks to its programmable neuro-core~\cite{Shrestha_etal24}.
In addition, it was shown that the main advantage of \ac{SNN} over ANN accelerators comes from exploiting the sparsity of spikes and not from the replacement of MAC by AC
operations~\cite{Dampfhoffer_etal23}.
Moving forward, graded spikes can expand the design space of neuromorphic computing algorithms beyond binary \acp{SNN} toward a broader category of \textit{sparse neural networks}, including but not limited to graded LIF and SigmaDelta neurons.
Further research on hardware and algorithms is required to explore this new paradigm of graded spikes, which can also refine the current abstraction level from biological neurons, moving beyond the binary spike representation to focus on sparsity.

Furthermore, we showed for the first time that we can deploy a \ac{SOTA} MCU13B spatial feature extractor on neuromorphic hardware, using patched inference to fit the model in a single Loihi~2 chip.
Coupled with the S4D temporal feature extractor, the model achieves the highest $F_1$ score of 83.6 \%, satisfies our real-time constraint with 16 predictions/s and only consumes 90~mW on Loihi~2.
Nevertheless, about 90\% of the total power is static power, which is the main limitation of the hardware system. 
Such static power can be reduced by removing unused components and manufacturing the next version of the Loihi chip with a more advanced technology process.
Our system serves as a proof-of-concept demonstrating the potential of embedding dedicated processing at the edge of \ac{EVS} via a direct interface, in which events are encoded as graded spikes and processed directly to fully leverage their spatio-temporal sparsity.
This approach enables low-latency and low-power perception at the system level.

Beyond the privacy-preserving aspect of edge computing, the always-on perception serves as a filter of raw sensory data by only sending out some data when a pattern of interest is detected (i.e., something \textit{relevant} happens). 
This drastically reduces the amount of data that need to be transferred, processed and stored in the cloud and thus reduces the overall energy consumption of the application ecosystem.
Neuromorphic technologies can enable this transition from cloud-centric to edge-distributed real-time processing toward a more balanced, efficient and scalable ecosystem, while leveraging the growth of IoT and edge sensing markets~\cite{YOLE_24,Muir_Sheik25}.
By starting from the application requirements and applying a system-level optimization through coherent sensor–processor–algorithm co-development, this work demonstrates a concrete step toward bringing neuromorphic technologies into the real world.

% -----------------------------------------------------------------------------------

\section*{Acknowledgment}
We thank Federico~Paredes-Vall\'es, Kirk~Scheper, and Sumit~Bam~Shrestha for their valuable feedback and support.

% -----------------------------------------------------------------------------------

% \clearpage

% \bibliographystyle{unsrt}
\bibliographystyle{IEEEtran}
\bibliography{biblio.bib}

\end{document}